\documentclass[conference]{IEEEtran}
\IEEEoverridecommandlockouts
\usepackage{cite}
\usepackage{amsmath,amssymb,amsfonts}
\usepackage{algorithmic}
\usepackage{booktabs}
\usepackage{multirow}
\usepackage{graphicx}
\usepackage{textcomp}
\usepackage{xcolor}
\usepackage{lipsum}
\usepackage{dblfloatfix}
\usepackage[T1]{fontenc}

\usepackage{caption}
\usepackage{tikz}
\usetikzlibrary{shapes,arrows}

\newcommand{\figref}[1]{Fig.~\ref{#1}}
\newcommand{\tblref}[1]{Table~\ref{#1}}
\newcommand{\x}{$\times$}
\graphicspath{{./figs/}}

\renewcommand{\baselinestretch}{0.96}

\begin{document}

\title{Extended Bit-Plane Compression for \\Convolutional Neural Network Accelerators
\thanks{The authors would like to thank \emph{armasuisse Science \& Technology} for funding this research. This project was supported in part by the EU's H2020 programme under grant no. 732631 (OPRECOMP).}
}
\author{
	\IEEEauthorblockN{Lukas Cavigelli, Luca Benini}
	\IEEEauthorblockA{Integrated Systems Laboratory, ETH Zurich, Switzerland --- \{cavigelli, benini\}@iis.ee.ethz.ch}
}

\maketitle

\begin{abstract}
After the tremendous success of convolutional neural networks in image classification, object detection, speech recognition, etc., there is now rising demand for deployment of these compute-intensive ML models on tightly power constrained embedded and mobile systems at low cost as well as for pushing the throughput in data centers. This has triggered a wave of research towards specialized hardware accelerators. Their performance is often constrained by I/O bandwidth and the energy consumption is dominated by I/O transfers to off-chip memory. 
We introduce and evaluate a novel, hardware-friendly compression scheme for the feature maps present within convolutional neural networks. We show that an average compression ratio of 4.4\x{} relative to uncompressed data and a gain of 60\% over existing method can be achieved for ResNet-34 with a compression block requiring <300\,bit of sequential cells and minimal combinational logic. \end{abstract}


\section{Introduction}
Computer vision has become a key ingredient for automatized data analysis over a board range of real-world applications: medical diagnostics \cite{Litjens2017}, industrial quality assurance \cite{Bian2016}, video surveillance \cite{Cavigelli2016a}, advanced driver assistance systems \cite{Wu2016b}, and many others. Many of these applications have only recently become feasible due to the tremendous increases in accuracy---even surpassing human capabilities \cite{HePReLU2015}---that have come with the rise of deep learning, and particularly, convolutional neural networks (CNNs, ConvNets). 

While CNN-based methods often require a significant computational effort, many of these applications should run in real-time on embedded and mobile systems. This has driven the development of specialized platforms, dedicated hardware accelerators, and optimized algorithms to reduce the number of compute operations as well as the precision requirements for the arithmetic operations \cite{Cavigelli2016,Cavigelli2015a,Albericio2016,Parashar2017,Zhang2016c,Aimar2017,Chen2016a,Han2016a,Cavigelli2018,Cavigelli2015}. 

When looking at these hardware platforms, the energy associated with loading and storing intermediate results/feature maps (and gradients during training) in external memory is not only significant, but often clearly higher than the energy used in computation and on-chip data buffering. This is even more striking when looking at networks optimized to reduce the computation energy by quantizing the weights to one bit, two bits, or power-of-two values, thereby eliminating the need for high-precision multiplications \cite{Andri2018,Courbariaux2015a,Andri2016,AojunZhou2016,Andri2016a}. 

Many compression methods for CNNs have been proposed over the last few years. However, many of them are focusing exclusively on 
\begin{enumerate}
 \item compressing the parameters/weights, which make up only a small share of the energy-intensive off-chip communication \cite{Chen2015a,Agustsson2017,Han2016}, 
 \item exploiting the sparsity of intermediate results, which is not always present (e.g. in partial results of a convolution layer or otherwise before the activation function is applied) and is not optimal in the sense that the non-uniform value distribution is not capitalized \cite{Rhu2018,Gudovskiy2018,Wang2017a}, 
 \item very complex methods requiring large dictionaries, or otherwise not suitable for a small, energy-efficient hardware implementation---often targeting efficient distribution and storage of trained models to mobile devices or the transmission of intermediate feature maps from/to mobile devices over a costly communication link \cite{Han2016}. 
\end{enumerate}

In this paper, we propose and evaluate a simple compression scheme for intermediate feature maps that can exploits sparsity as well as the distribution of the remaining values. It is suitable for a very small and energy-efficient implementation in hardware (<300\,bit of registers), and could be inserted as a stream (de-)compressor before/after a DMA controller to compress the data by 4.4\x{} for 8\,bit AlexNet.

\section{Related Work}
There are several methods out there describing hardware accelerators which exploit feature map sparsity to reduce computation: Cnvlutin \cite{Albericio2016}, SCNN \cite{Parashar2017}, Cambricon-X \cite{Zhang2016c}, NullHop \cite{Aimar2017}, Eyeriss \cite{Chen2016a}, EIE \cite{Han2016a}. Their focus is on power gating or skipping some of the operations and memory accesses. While this automatically entails defining a scheme to feed the data into the system, minimizing the bandwidth was not the primary objective of any of them. They all use one of three methods: 
\begin{enumerate}
 \item Zero-RLE (used in SCNN): A simple run-length encoding for the zero values, i.e. a single prefix bit followed by the number of zero-values or the non-zero value. 
 \item Zero-free neuron array format (ZFNAf) (used in Cnvlutin): Similarly to the widely-used compressed sparse row (CSR) format, non-zero elements are encoded with an offset and their value. 
 \item Compressed column storage (CCS) format (e.g. used in EIE): Similar to ZFNAf, but the offsets are stored in relative form, thus requiring less bits to store them. Few bits are sufficient, and in case they are all exhausted, a zero-value can be encoded as if it was non-zero. 
\end{enumerate}
\begin{figure*}[b]
	\includegraphics[width=\linewidth]{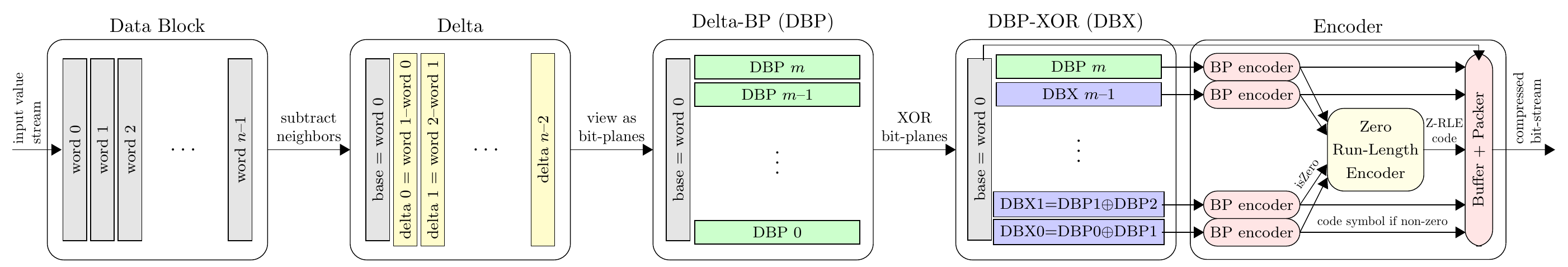}
	\caption{Overview of the processing steps to apply bit-plane compression.}
    \label{fig:BPCoverview}
\end{figure*}

Most compression methods are focusing on minimizing the model size. Most of them are very complex (area) to implement in hardware and need large dictionaries. 
One such method, deep compression \cite{Han2016}, combines pruning, trained clustering-based quantization, and Huffman coding. Most of these steps involved cannot be applied to the intermediate feature map, which change for every inference as opposed to the weights which are static and can be optimized off-line. Furthermore, applying Huffman coding---while being optimal---implies storing a large dictionary (typically several MB).
Similar issues arise when using Lempel-Ziv-Welch (LZW) coding \cite{Welch1984,Ziv1978} as present in e.g. the ZIP compression scheme, where the dictionary is encoded in the compressed data stream. This makes it unsuitable for a lightweight and energy-efficient VLSI implementation \cite{Lin2006,Zhou2016}. 

Few more methods exist by changing the CNN's structure in order to compress the weights \cite{Chen2015a,Agustsson2017} or the feature maps \cite{Gudovskiy2018,Wang2017a}. However, they require altering the CNN's model and retraining, and they introduce some accuracy loss. Furthermore, they can only be used to compress a few feature maps at specific points within the network and introduce additional compute effort, such as applying a Fourier transform to the feature maps. 

The most directly comparable approach, cDMA \cite{Rhu2018}, describes a hardware-friendly compression scheme to reduce the data size of intermediate feature maps. Their target application differs in that their main goal is to allow faster offloading of the feature maps from GPU to CPU memory through the PCIe bandwidth bottleneck during training, thereby enabling larger batch sizes and deeper and wider networks without sacrificing performance. They propose to use \emph{zero-value compression (ZVC)}, which takes a block of 32 activation values, and generates a 32-bit mask where only the bits to the non-zero values are set. The non-zero values are transferred after the mask. This provides the main advantage over Zero-RLE that the resulting data volume is independent of how the values of the feature maps are serialized while also providing small compression ratio advantages. Note that this is a special case of Zero-RLE with a maximum zero burst length of 1.

For this work, we build on a method known in the area of texture compression for GPUs, \emph{bit-plane compression (BPC)} \cite{Kim2016}, fuse it with sparsity-focused compression methods, and evaluate the resulting compression algorithm on intermediate feature maps to show compression ratios of 4.4\x{} and 2.8\x{} for 8\,bit AlexNet and SqueezeNet, respectively. 

\section{Compression Algorithm}
\begin{figure}
	\includegraphics[width=\linewidth]{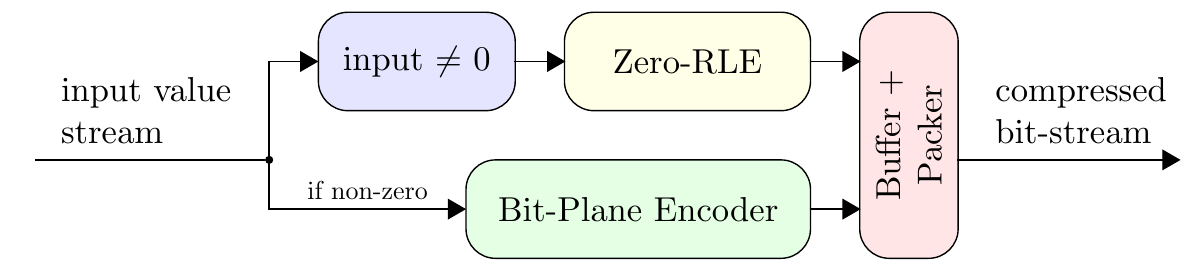}
	\caption{Top-level view of the proposed compression scheme.}
    \label{fig:comprOverview}
\end{figure}

An overview of the proposed algorithm is shown in \figref{fig:comprOverview}. The value stream is decomposed into a zero/non-zero stream on which we apply run-length encoding to compress the zero burst commonly occurring in the data, and a stream of non-zero values which we encode using bit-plane compression. The later compresses a fixed number of words $n$ jointly, and the resulting compressed bit-stream is injected immediately after at least $n$ non-zero values have been compressed.

\subsection{Zero/Non-Zero Encoding with RLE}
The run-length encoder simply compresses bursts of 0s with a single 0 followed by a fixed number of bits which encode the burst length. Non-zero values, at this point 1-bits, are not run-length encoded, i.e. for each of them a 1 is emitted. If the length of a zero-burst exceeds the corresponding maximum burst length, the maximum is encoded and the remaining bits are encoded independently, i.e. in the next code symbol. 

\subsection{Bit-Plane Compression}
An overview of the bit-plane compressor (BPC) used to compress the non-zero values is shown in \figref{fig:BPCoverview}. For BPC a set of $n$ words of $m$\,bit, a \emph{data block}, is compressed by first building differences between each two consecutive words and storing the first word as the base. This exploits that neighboring values are often similar. 

The data items storing these differences are then viewed as $m+1$ bit-planes of $n$\,bit each (delta bit-planes, DBPs). Neighboring DBPs are XOR-ed, now called \emph{DBX}, and the DBP of the most significant bit is kept as the base-DBP. The results are fed into bit-plane encoders, which compress the DBX and DBP values to a bit-stream following \tblref{tbl:bitplaneSymbolEnc}. Most of these encodings are applied independently per DBX symbol. However, the first can be used to jointly encode multiple consecutive bit-planes at once, if they are all zero. This is where the correlation of neighboring values is best exploited. Note also the importance of the XOR-ing step in order to map two's complement negative values close to zero also to words consisting mostly of zero-bits. 
\begin{table}
  \centering
  \caption{Bit-Plane Symbol Encoder}
  \label{tbl:bitplaneSymbolEnc}
  \begin{tabular}{lrl}
  	\toprule
  	DBX Pattern & Length [bit] & Code (binary) \\ 
  	\midrule
  	0 (run length 2 to m+1) & $2+\log_2 m$ & 001 \& to\_bin(runLength-2) \\
  	0 (run length 1) & 3 & 01 \\
  	All-1 & 5 & 00000 \\
  	DBX!=0 \&\& DBP=0 & 5 & 00001 \\
  	Two consecutive 1s & $5+\log_2 m$ & 00010 \& to\_bin(posOfFirstOne) \\
  	Single 1 & $5+\log_2 m$ & 00011 \& to\_bin(posOfOne) \\
  	Uncompressed 1 & $1+m$ & 1 \& to\_bin(DBX word) \\
  	\bottomrule
  \end{tabular}
\end{table}
The proposed compression method can be applied to integers of various word widths, but also to floating-point data types, although this affects the compression ratio negatively.

\subsection{Hardware Suitability}
\begin{figure}
	\includegraphics[width=\linewidth]{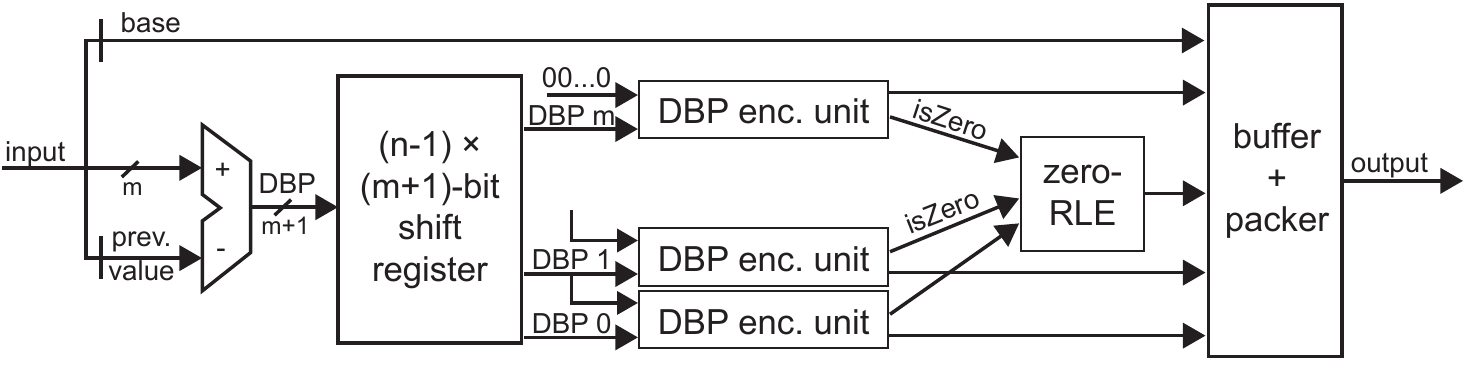}
	\caption{Block diagram of the bit-plane encoder.}
    \label{fig:bpEncBlockDiag}
\end{figure}
The proposed algorithm is very hardware friendly: no code-book needs to be stored, just a few data words need to be kept in memory. From the overview (cf. \figref{fig:comprOverview}), the Zero-RLE mostly consists of a counter and the non-zero check is also negligible in size. The buffer and packer assembles the bit-stream and needs very little logic and a few bits of storage to pack the resulting bit-stream into words. The last remaining unit, the bit-plane encoder, is shown in \figref{fig:bpEncBlockDiag}. In terms of registers, only the \emph{base} value ($m$\,bit), the previous value to build the differences ($m$\,bit), and a $(n-1)\times (m+1)$\,bit shift register are needed (with e.g. $n=m=16$ a total of $<300$\,bit). Only very little logic is required as well: a single subtractor, a simple zero-RLE encoder, and the DBP encoder unit realizing the mapping in \tblref{tbl:bitplaneSymbolEnc}. 

Also the logic operations are very regular and fairly low-cost in terms of size and energy. The resulting compression reduces the energy spent on interfaces to DRAM, on inter-chip or back-plane communication---the corresponding standards specify very efficient power-down modes \cite{lpddr4,gddr5}---as well as potentially saving DRAM refresh cycles for the saved memory area \cite{Litjens2017}, and providing an alternative to increasing the bandwidth of such interfaces, which would imply more expensive packages, circuit boards, and additional on-chip circuits (e.g. PLLs, on-chip termination, etc.) \cite{lpddr4,gddr5}. 

\section{Results}

\subsection{Experimental Setup}
Where not otherwise stated, we perform our experiments on AlexNet and are using images from the ILSVRC validation set. All models we used were pre-trained and downloaded from the PyTorch/Torchvision data repository. Some of the experiments are performed with fixed-point data types (default: 16-bit fixed-point). For these, the feature maps were normalized to exploit the full range, i.e. the worst-case scenario from a compression point of view. All the feature maps were extracted after the ReLU activations. 

\subsection{Sparsity, Activation Histogram \& Data Layout}
\begin{figure}
	\centering
	\includegraphics[width=\linewidth]{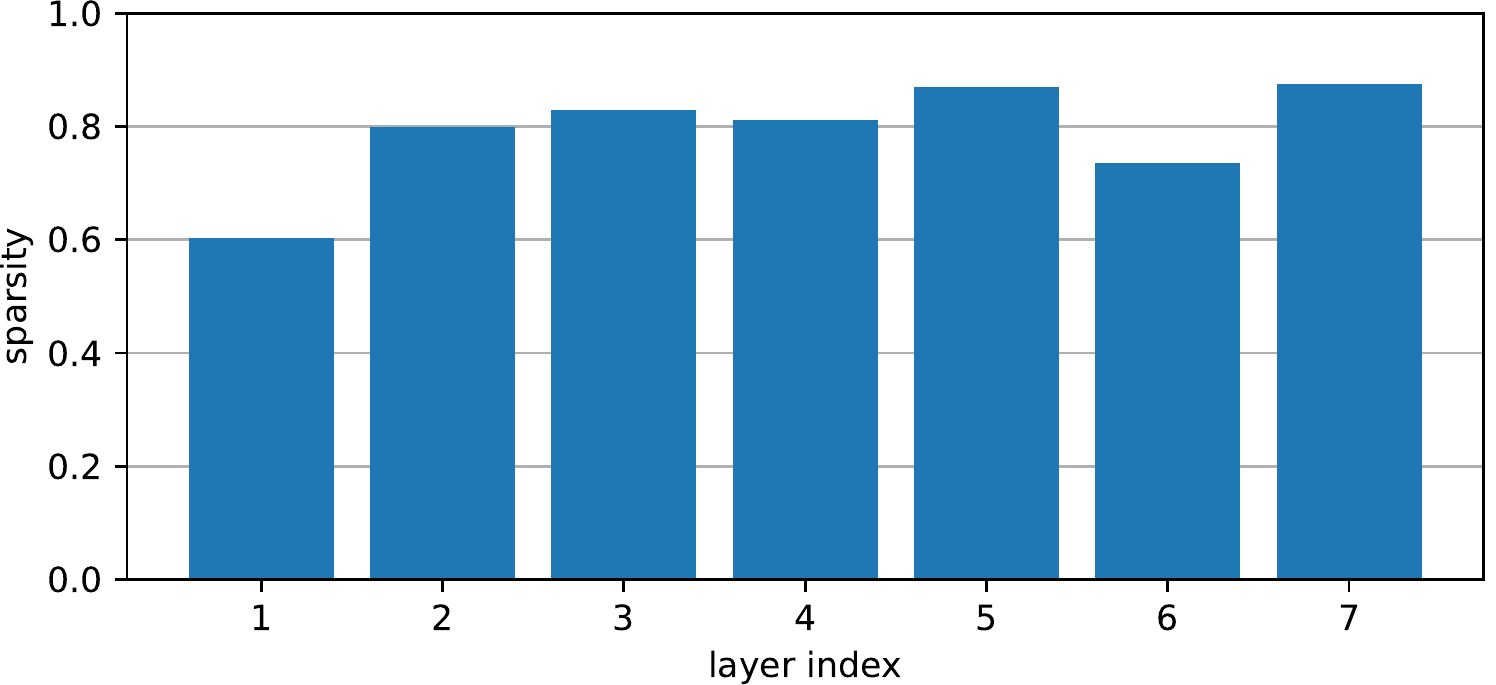}
	\caption{Feature map sparsity after activation for each layer in AlexNet.}
	\label{fig:sparsity}
\end{figure}

Neural networks are known to have sparse feature maps after applying a ReLU activation layer, which can be applied on-the-fly after the convolution layer and possibly batch normalization. However, it varies significantly for different layers within the network as well as for different CNNs. Sparsity is a key aspect when compressing feature maps, and we analyze it in \figref{fig:sparsity}.

\begin{figure}
	\centering
	\includegraphics[width=\linewidth]{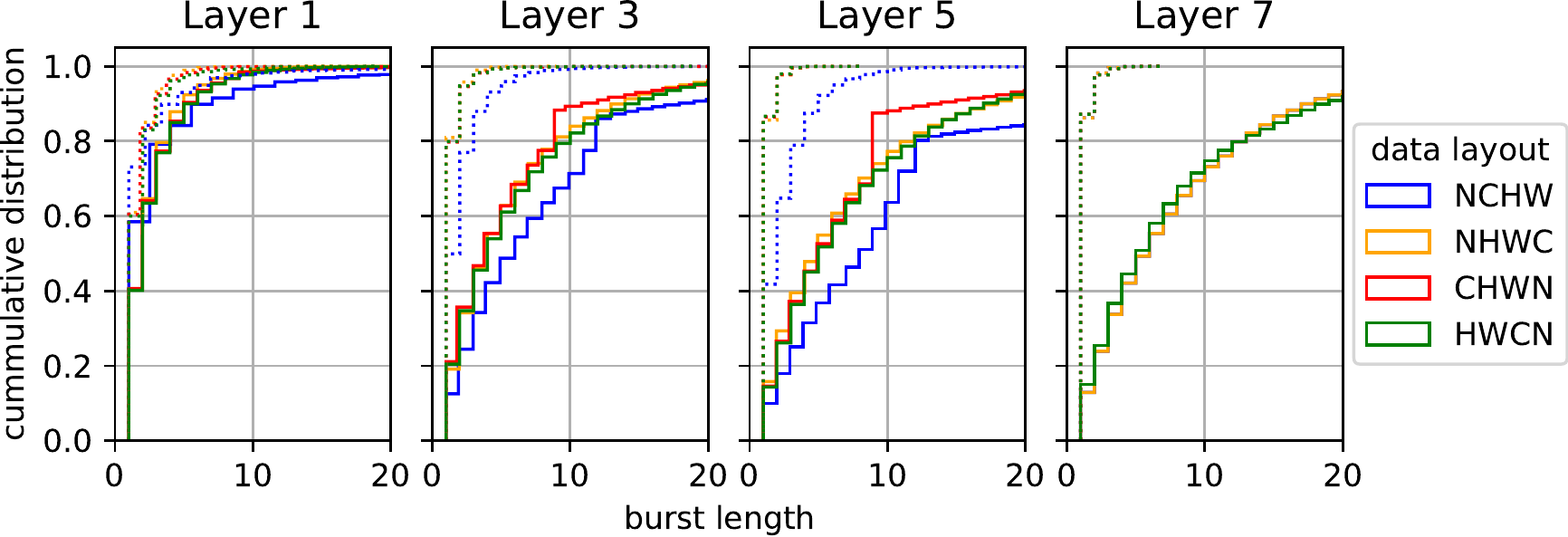}
	\caption{Cumulative probability distribution of zero (solid) and non-zero (dotted) burst lengths.}
	\label{fig:zeroBurstLens}
\end{figure}

The sparse values are not independently distributed but rather occur in bursts when the 4D data tensor is laid out in one of the obvious formats. The most commonly used formats are NCHW and NHWC, which are those supported by most frameworks and the widely used Nvidia cuDNN backend. NCHW is the preferred format for cuDNN and the default memory layout and means that neighboring values in horizontal direction are stored next to each other in memory before the vertical, channel, and batch dimensions. NHWC is the default format of TensorFlow and has long before been used in compute vision and has the advantage of simple non-strided computation of inner products in channel (i.e. feature map) dimension. Further reasonable options which we include in out analysis are CHWN and HWCN, although most use-cases with hardware acceleration are targeting real-time low-latency inference and are thus operating with a batch size of 1. We analyze the distribution of the length of zero bursts for the these four data layouts at various depths within the network in \figref{fig:zeroBurstLens}. 

The results clearly show that having the spatial dimensions (H, W) in next to each other in the data stream provide the longest zero bursts (lowest cumulative distribution curve) and thus the better compressibility than the other formats. This is also aligned with intuition: feature maps values mark the presence of certain features and can be expected to be smooth. Inspection the feature maps of CNNs is commonly known to show that they behave like 'heat maps' marking the presence of certain geometric features nearby. Based on these results we perform all the following evaluations based on the NCHW data layout. Not also that the burst length of non-zero values is mostly very short, such that there is limited gain in applying RLE also for the one-bits.

\begin{figure}
	\centering
	\includegraphics[width=\linewidth]{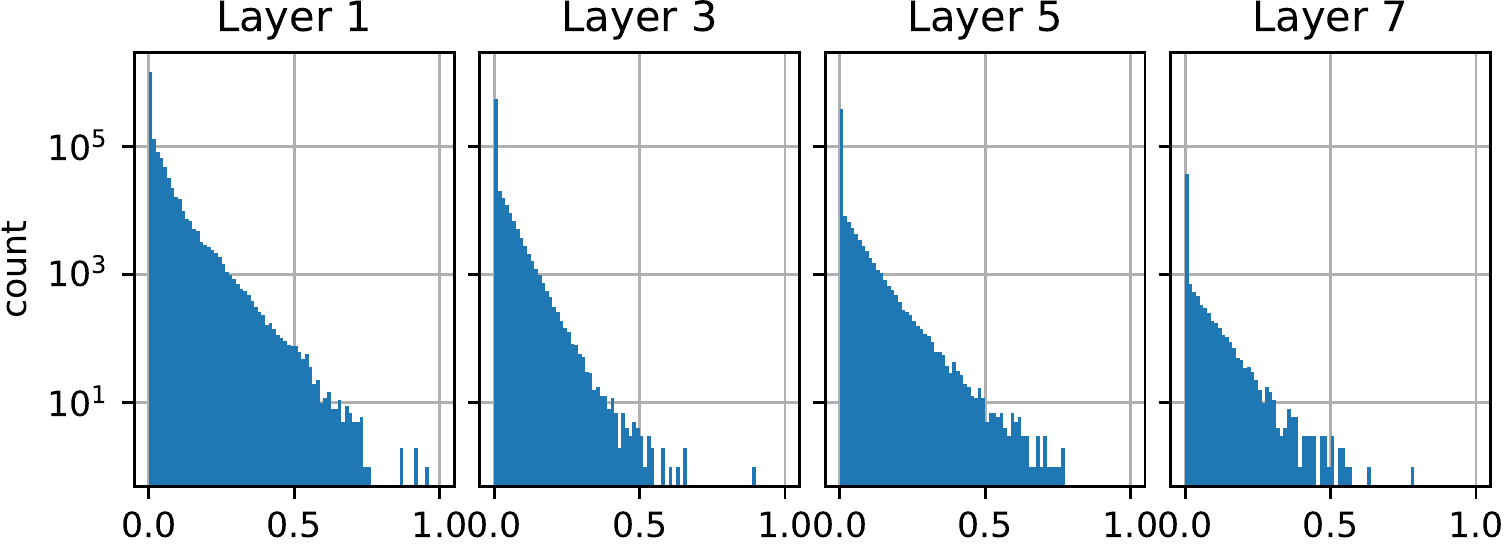}
	\caption{Histogram of activation values at various depths within the network. Note the logarithmic vertical axis. }
	\label{fig:hist}
\end{figure}

To compress further beyond exploiting the sparsity, the data has to remain compressible. This is definitely the case as can be seen when looking at histograms of the activation distributions as shown in \figref{fig:hist} and a strong indication that additional compression of the non-zero data is possible. 

\subsection{Selecting Parameters}
The proposed method has two parameters: the maximum length of a zero sequence that can be encoded with a single code symbol of the Zero-RLE, and the BPC block size ($n$, number of non-zero word encoded jointly). 

\paragraph*{Max. Zero Burst Length}
We first analyze the effect of varying the maximum zero burst length for Zero-RLE on the compression ratio without for various data wordwidths in \tblref{tbl:zeroRLEburstLen}. 
\begin{table}
	\centering
	\caption{Compression Ratio Using ZVC And Zero-RLE for Various Maximum Zero Burst Lengths}
	\label{tbl:zeroRLEburstLen}
	\begin{tabular}{r|r|rrrrrr}
	\toprule
		\multirow{2}{*}{wordwidth} & \multirow{2}{*}{ZVC} & \multicolumn{6}{c}{Zero-RLE max. zero burst length} \\
	    &    &   $2^1$ &   $2^2$ &   $2^3$ &   $2^4$ &   $2^5$ &   $2^6$ \\
	\midrule
	  8 &  2.52 &  2.48 &  2.56 &  2.62 &  \textbf{2.63} &  2.59 &  2.53 \\
	 16 &  3.00 &  2.96 &  3.02 &  3.06 &  \textbf{3.07} &  3.04 &  3.00 \\
	 32 &  3.30 &  3.28 &  3.32 &  3.34 &  \textbf{3.35} &  3.33 &  3.31 \\
	\bottomrule
	\end{tabular}
\end{table}
The optimal value is arguably the same for our proposed method, since an constant offset in compressing the non-zero values does not affect the optimal choice of this parameter (just like to wordwidth has no effect on it). The results also serve as a baseline for Zero-RLE and ZVC. It is worth noting that ZVC corresponds to Zero-RLE with a max. burst length of 1, yet breaks the trend shown in \tblref{tbl:zeroRLEburstLen}. This is due to an inefficiency of Zero-RLE in this corner: for a zero burst length of 1, ZVC requires 1\,bit whereas Zero-RLE with a max. burst length of 2 takes 2\,bit. For a zero burst of length 2, ZVC encode 2 symbols of 1\,bit each and Zero-RLE takes 2\,bit as well. ZVC thus always performs at least as well for such a short max. burst length. 

\begin{figure}
	\centering
	\includegraphics[width=\linewidth]{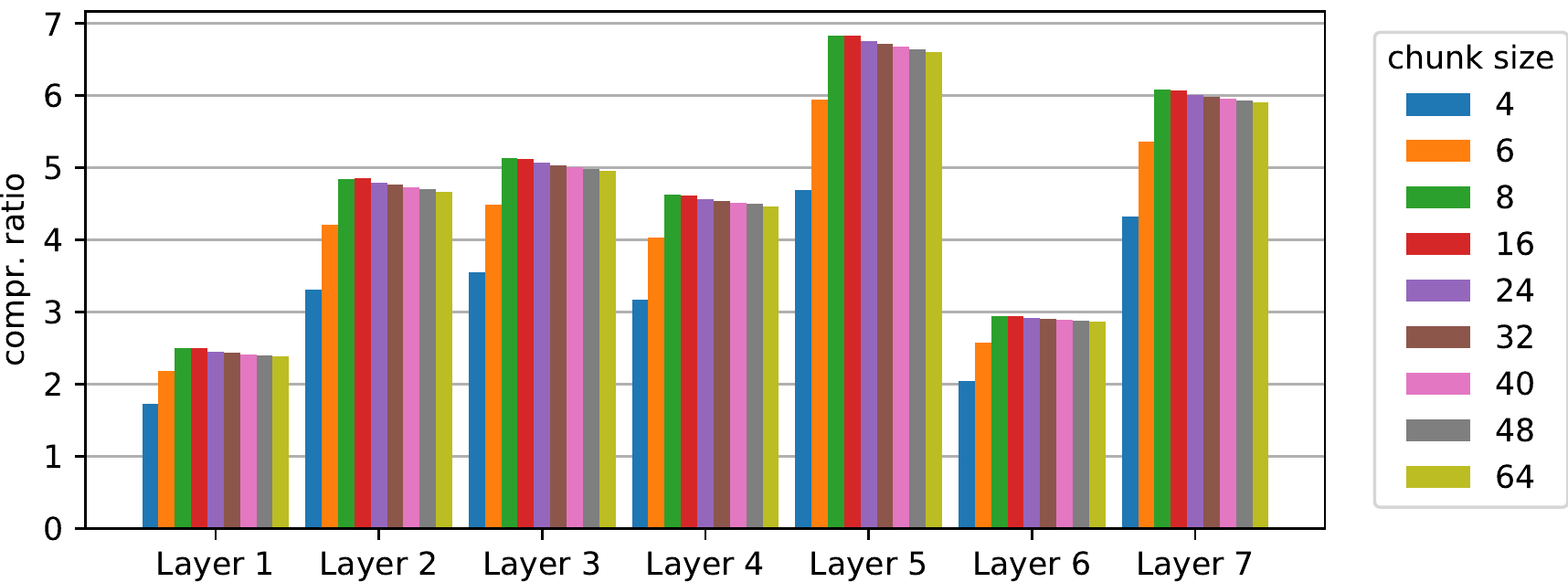}
	\caption{Analysis of the compression ratios of various layers' outputs for several BPC block sizes and with 16-bit fixed-point values.}
	\label{fig:chunkSizeEval}
\end{figure}

\paragraph*{BPC Block Size}
We analyze the effect of the BPC block size parameter in \figref{fig:chunkSizeEval} at various depths within the network. The best compression ratio is achieved with a block size of 16 across all the layers. A block size of 8 might also be considered to minimize the resources of the (de-)compression hardware block at a small drop in compression ratio. 

\subsection{Total Compression Factor}
\begin{figure}
	\centering
	\includegraphics[width=\linewidth]{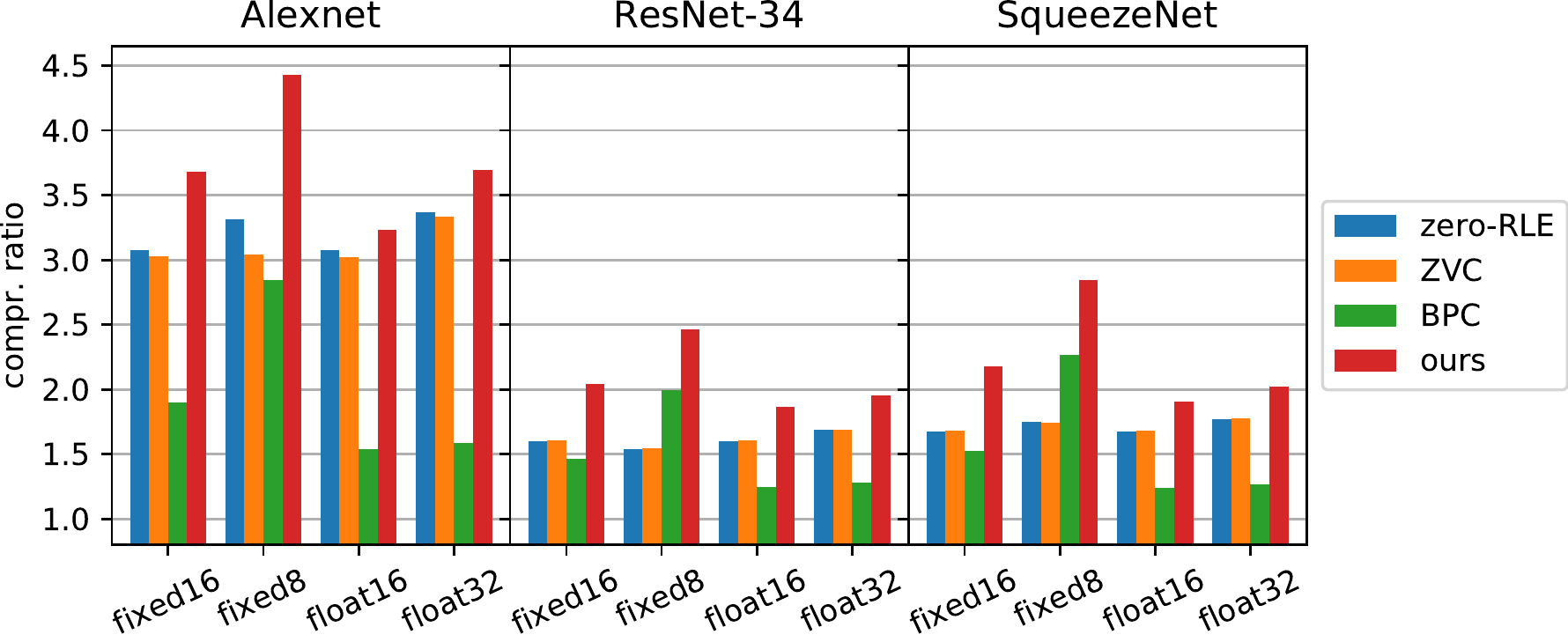}
	\caption{Evaluation of the compression ratio of all feature maps of AlexNet, ResNet-34, and SqueezeNet for various compression methods and data types.}
	\label{fig:totalComprRate}
\end{figure}

We analyze the total compression factor of all feature maps of AlexNet, ResNet-34, and SqueezeNet in \figref{fig:totalComprRate}. For AlexNet, we can notice the high compression ratio of around 3\x{} already introduced by Zero-RLE and ZVC and that it is very similar for all data types. We further see that pure BPC is not suitable since it introduces too much overhead when encoding only zero-values. For ResNet-34 and SqueezeNet, the gains by exploiting only the sparsity is significantly lower at around 1.55\x{} and 1.7\x{}. The proposed method outperforms previous approaches clearly with compression ratios of 4.45\x{}, 2.45\x{}, and 2.8\x{} (for 8-bit fixed-point), respectively. 

The gains for 8-bit fixed-point data is significantly higher than for other data formats. Most input data---also CNN feature maps---carry the most important information is in the more significant bits and in case of floats in the exponent. The less significant bits appear mostly as noise to the encoder and cannot be compressed without accuracy loss, such that this behavior---a lower compression ratio for wider word widths---is expected. 

\section{Conclusion}
We have presented and evaluated a novel compression method for CNN feature maps. The proposed algorithm achieves an average compression ratio of 4.4\x{} on AlexNet (+35\% over previous methods), 2.45\x{} on ResNet-34 (+60\%), and 2.8\x{} on SqueezeNet (+65\%) for 8\,bit data, and thus clearly outperforms state-of-the-art, while fitting a very tight hardware resource budget with <300\,bit of data and very little compute logic.

\newpage
\bibliographystyle{IEEEtran}
\bibliography{library}

\end{document}